\documentclass[letterpaper, 10 pt, conference]{ieeeconf}  

\IEEEoverridecommandlockouts                              

\usepackage{graphics} 
\usepackage{epsfig} 
\usepackage{mathptmx} 
\usepackage{times} 
\usepackage{amsmath} 
\usepackage{amssymb}  
\usepackage{multirow}
\usepackage{multicol}
\usepackage{hyperref}
\usepackage{makecell}
\usepackage{textgreek}
\usepackage{forest}

\title{\LARGE \bf
Brno Urban Dataset - The New Data for \\ Self-Driving Agents and Mapping Tasks
}

\author{Adam Ligocki$^{1}$, Ales Jelinek$^{1}$ and Ludek Zalud$^{1}$
\thanks{The research was supported by ECSEL JU under the project H2020 737469 AutoDrive - Advancing fail-aware, fail-safe, and fail-operational electronic components, systems, and architectures for fully automated driving to make future mobility safer, affordable, and end-user acceptable. This research has been financially supported by the Ministry of Education, Youth and Sports of the Czech republic under the project CEITEC 2020 (LQ1601).}
\thanks{$^{1}$All the authors are with the Central European Institute of Technology (CEITEC), 
        Cybernetics in Material Science research group, 
        Brno University of Technology, Purkynova 123, Brno-Kralovo Pole, Czechia,
        {\tt\small adam.ligocki@ceitec.vutbr.cz}, 
        {\tt\small ales.jelinek@ceitec.vutbr.cz}, 
        {\tt\small ludek.zalud@ceitec.vutbr.cz}}
}

\begin{document}

\bstctlcite{BSTcontrol}

\maketitle
\thispagestyle{empty}
\pagestyle{empty}

\begin{abstract}
Autonomous driving is a dynamically growing field of research, where quality and amount of experimental data is critical. Although several rich datasets are available these days, the demands of researchers and technical possibilities are evolving. Through this paper, we bring a new dataset recorded in Brno - Czech Republic. It offers data from four WUXGA cameras, two 3D LiDARs, inertial measurement unit, infrared camera and especially differential RTK GNSS receiver with centimetre accuracy which, to the best knowledge of the authors, is not available from any other public dataset so far. In addition, all the data are precisely timestamped with sub-millisecond precision to allow wider range of applications. At the time of publishing of this paper, recordings of more than 350~km of rides in varying environment are shared at: \url{https://github.com/RoboticsBUT/Brno-Urban-Dataset}.
\end{abstract}

\section{Introduction}

Research in the domain of autonomous mobile vehicles have tremendously expanded in the last few years \cite{Badue2019}, \cite{Faisal2019}, \cite{Litman2019}. From one of many possible applications of general mobile robotics \cite{Cadena2016} and a geeky interest of technical visionaries it became a large topic for both scientific and commercial sectors. Despite the undoubted motivation of financial bounties and pursuit of the emerging trends, this boom is also fueled with openly available data allowing more people to be part of it. To equip a car with state of the art sensors can easily become too expensive for small subjects such as start-ups or research groups on local universities. Sharing data allows much more researchers to participate in the progress of the field and enrich it with novel ideas, which, in the end, rewards everybody \cite{McKiernan2016}. Second good reason for data sharing is a possibility to bypass the necessity of building and maintaining the sensory apparatus, which otherwise requires extra resources and engineering skills not related to the actual research topic of artificial intelligence. Having the opportunity to build our own data acquisition system and exceeding current state of the art in some of its parameters, we have decided to make the data publicly available.

\begin{figure}[t]
    \centering
    \includegraphics[width=8cm]{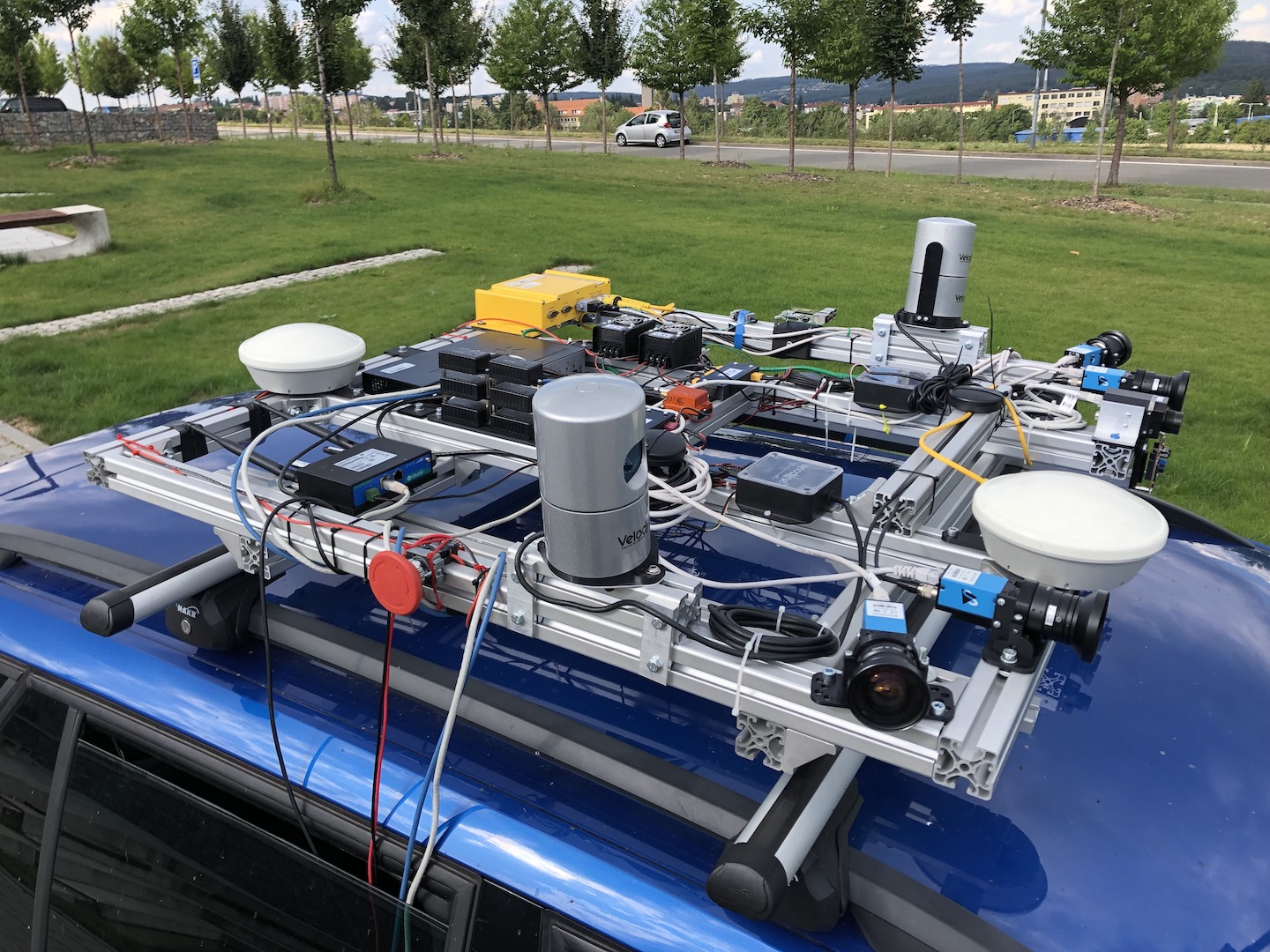}
    \caption{The detail of the Atlas sensory framework, which has been used to record all the data published in the Brno Urban Dataset. On the image there are four RGB cameras, single thermal (IR) camera, two Velodyne \mbox{HDL-32E} LiDARs, the RTK GNSS receiver with pair of differential antennas and the IMU unit in the center of the frame.}
    \label{fig_preview}
\end{figure}

Through this paper, we provide an urban dataset recorded in Brno - Czech Republic and its near surroundings. The location offers highly diverse environments - from highways to farm roads, from densely build-up areas to woods, all recorded in conditions of the real traffic. The sensory system contains all sensors which are considered standard these days \cite{Badue2019}, \cite{Yin2017}: frontal stereo cameras, lateral cameras, two 3D laser scanners (LiDARs), inertial measurement unit (IMU) and global navigation satellite system (GNSS) receiver. Beyond that, there are three key features, which make the system unique. First, to our best knowledge, it is the only measurement system equipped with differential real-time kinematic (RTK) GNSS receiver providing centimetre accurate global positioning and heading at the same time. Second, a thermal camera is used to sense the scene in front of the car, which greatly enhances vulnerable road user awareness and general detection of various objects in bad weather conditions. Third, most of the system is exactly synchronized and timestamped using the GPS signal with sub-millisecond precision. 

Usage of such data is wide. From practical point of view, having a reliable reference eliminates many problems in data processing. In general pose estimation, the benefits of differential GNSS are well understood \cite{Schneider2016}, \cite{Schall2009} and its application in autonomous road vehicles seems reasonable. Similarly, issues arising from bad synchronization of cameras are subject of dedicated research coming-up with creative ways of dealing with them \cite{Albl2017a}, \cite{Mhiri2014}, \cite{Svedman2005}, but our system allows to avoid such problems entirely, which is probably going to be the case in modern autonomous vehicles as well. 

Another possible application of our data consists in development of the previously mentioned methods for dealing with uncertainty, because in some occasions their adoption is necessary. Exact timestamping allows evaluation of such methods, where timing of the data acquisition is part of the stochastic model \cite{Levinson2016}, \cite{LeGentil2018}. With precise global localization, we can benchmark the methods relying on local sensors such as cameras \cite{Mhiri2014}, \cite{Cortes2018}, LiDARs \cite{Wang2017a}, \cite{Wolcott2014} and their combinations, frequently enhanced with inertial measurement units \cite{Deilamsalehy2017}. As we will show in the next section, the purpose of most of the other datasets is different and would be of limited usage in such experiments.

\section{Related Works}
\label{sec_sota}

Publicly available datasets are rapidly evolving according to the fast technological progress of the last two decades. Sensor accuracy and especially resolutions have grown substantially, which led to larger memory requirements for storing and higher bandwidth needs for sharing of the recordings. Today's possibilities fulfill all these demands, which makes older datasets exceeded quite fast. For example, going through the review in the M\' alaga dataset paper \cite{Blanco-Claraco2014}, we see that datasets captured five to ten year ago provide mostly 2D laser scans, camera resolution of a fraction of a megapixel and the largest had tens of kilometers of recorded path. Comparing them with more recent surveys \cite{Yin2017} and \cite{Badue2019}, the standard have risen significantly. Unfortunately many large datasets are not freely available \cite{Wang2017}. The following paragraphs cover only the state of the art represented by those opened to public interest.

\subsection{Special Purpose Datasets}
The recherche would be incomplete without a note on specialized datasets for distinct tasks in automated driving. Long solved is the traffic sign recognition problem \cite{Mogelmose2012} with dedicated datasets such as \cite{Stallkamp2012}, \cite{Timofte2014}. Similar is the traffic light detection and interpretation \cite{Fregin2018}. The other datasets are more focused on mobile devices and augmented reality, therefore they do not cover the traffic from the perspective of the vehicle, but rather of a pedestrian as e.g. \cite{Cortes2018}. Another distinguishable group are the synthetic datasets such as SYNTHIA \cite{Ros2016} or P.F.B. \cite{Richter2017}, which are specific due to precisely computed yet somewhat simplifying output data. The WildDash dataset \cite{Zendel2018}, is focused on data, where the image segmentation algorithms fail. Contrary to its larger counterparts, this dataset is not meant for learning of the algorithms, but mainly for testing of their results.

\subsection{Vision-focused datasets}
Many datasets are designed to serve mostly for machine vision research, especially segmentation of real-life scenes and recognition of objects of interest. Large amount of very short recordings with a few manually annotated frames and crude GPS positioning is usually all what is available. The key contribution consists in large variety of scenes (often acquired though crowd sourcing) and reliable reference data. The CityScapes dataset \cite{Cordts2016}, Mapillary Vistas dataset \cite{Neuhold2017} and the Berkeley DeepDrive dataset \cite{Yu2018} are the main examples. 

\subsection{General mapping datasets}
Arguably the most important group of datasets strongly focuses on sensory quality and variety, which allows large amount of applications in research. The vision subsystems are usually of higher quality and accompanied with laser scanners and inertial units. The most notable these days are the KITTI \cite{Geiger2013}, the M\' alaga urban \cite{Blanco-Claraco2014}, the Oxford RobotCar \cite{Maddern2017} and the ApolloScape \cite{Huang2018} datasets summarized in Tab.~\ref{tab_datasets}.

\begin{table}[t]
\centering
\begin{tabular}{c|c|c|c}
 \hline
 Dataset & Cameras & LiDARs & GNSS \\
 \hline
 \makecell{KITTI} & 4x 1.4~MP, 10~FPS & 2000x64~pt, 10~Hz & RTK GPS \\
 \hline
 M\' alaga & 2x 0.8~MP, 20~FPS & \makecell{2x 1080~pt, 75~Hz \\ 3x 1080~pt, 40~Hz} & GPS  \\
 \hline
 Oxford & \makecell{3x 1.2~MP, 16~FPS\\3x 1~MP, 11~FPS} & \makecell{2x 540~pt, 50~Hz \\ 680x4~pt, 12.5~Hz} & GPS \\
 \hline
 ApolloScape & 4x 9.2~MP, 30 FPS & 2x 1000000~pt/s & RTK GPS \\
 \hline
\end{tabular}
\caption{An overview of the datasets for general mapping.}
\label{tab_datasets}
\end{table}

The size of the datsets varies a lot, the KITTI and M\' alaga recordings cover only tens of kilometers, while the Oxford and the ApolloScape sets map 1000+ kilometers of roads. Treatment of timing and synchronization is unique in each case, the Oxford dataset timing has an excellent sub-millisecond precision, authors of the ApolloScape set claim just synchronization with no details explained and the KITTI and M\' alaga provide less precisely timed and mostly not synchronized data.

\subsection{Summary}
The previous overview covers many datasets relevant to autonomous driving, but only a few are of the kind, which we are presenting. Removing the vision only datasets \cite{Cordts2016}, \cite{Neuhold2017}, \cite{Yu2018} and \cite{Zendel2018}, we are left with four projects providing similar data as we do. The KITTI \cite{Geiger2013}, M\' alaga \cite{Blanco-Claraco2014} and Oxford \cite{Maddern2017} datasets offer lower quality sensory data, while the ApolloScape \cite{Huang2018} is clearly the richest dataset available in sense of camera resolution and point cloud density. On the other hand, none of these four references contain differential RTK GPS and, although synchronization and timestamping is mentioned everywhere, only the Oxford dataset presents detailed treatment of the topic using special software. We have achieved the same accuracy with more stable results due to hardware precautions. Although we offer only raw data, taking into account current state of the art, the Brno urban dataset has features reaching beyond that margin.

\section{The Data Acquisition Platform}

The dataset has been recorded using an extensible sensory platform called ATLAS built in our laboratory. The system is composed of a communication network, data processing computer, synchronization unit and sensors themselves. The whole apparatus was designed with precision and modularity in mind, allowing wide range of experimental setups, while maintaining quality of the recordings. The following paragraphs will go through the current state of the ATLAS platform (see Fig.~\ref{fig_preview}) and the Conclusion will summarize the upcoming extensions.

\subsection{Data Gathering Infrastructure}

Because we deal with large throughput of raw data, the central control and recording computer is built on a CPU with 64 PCIe lanes and a NMVe SSD disk. A Nvidia GTX 1080Ti graphic card is present as well for video processing and compression. The backbone of the ATLAS recording framework is an Ethernet network. It is based on IP communication protocol and all the sensors and the acquisition PC are interconnected via a high-speed switch with up to 18~Gb/s bandwidth for connection to the acquisition PC. The only exception is the IMU which is connected to the PC through a virtual serial link via USB interface. 

It should be noted, that neither the recording PC, nor the Ethernet network exhibit real-time capabilities for precise and reliable timing and this functionality is solved independently as will be discussed in Sec.~\ref{sec_syn_and_stamp}.

\subsection{Sensory Equipment}








\begin{table}[t]
\centering
\begin{tabular}{c|c|c|c|c}
 \hline
 Sensor & Type & Details & Freq. & Output data\\
 \hline
 \makecell{4x RGB \\ camera} & \makecell{DFK33- \\ -GX174} & \makecell{$75^\circ$ FoV \\ 1920x1200~px} & 10 & h265 video \\
 \hline
 \makecell{Thermal \\ camera} & FLIR Tau 2 & \makecell{$70^\circ$ FoV \\ 640x512~px} & 30 & h265 video \\
 \hline
 \makecell{2x LiDAR} & \makecell{Velodyne\\HDL-32e} & \makecell{32 beams \\ $\sim$2000 pts/turn} & 10 &  point cloud \\
 \hline
 \makecell{GNSS \\ receiver} & \makecell{Trimble \\ BX982} & \makecell{RTK accuracy \\ Direct heading} & \makecell{ 20 \\ 20 } & \makecell{global pose \\ time} \\
 \hline

 \makecell{IMU} & \makecell{Xsens \\ MTi-G-710} & \makecell{Combined \\ non-electrical \\ quantities \\ sensor} & \makecell{400\\400\\100\\400\\400\\400\\50} & \makecell{ accelerometer \\ gyroscope \\ magnetometer \\ temperature \\ global pose \\ time\\ pressure} \\

 \hline
\end{tabular}
\caption{An overview of the sensors installed on the ATLAS measurement platform.}
\label{tab_sensors}
\end{table}

Sensors are the clearly a crucial part of the system. Table~\ref{tab_sensors} summarizes all the devices currently installed and their most important parameters. In the next paragraphs, we will briefly elaborate on each sensor category to better present the data we provide.

As seen in Sec.~\ref{sec_sota}, RGB cameras are a must in autonomous vehicle applications. The ATLAS platform has two cameras installed for stereo vision in the front and two lateral cameras with wider field of view (FoV) for better coverage of the crossroads, walkways and other road users passing around. The frontal cameras are installed wide apart ($\sim$70~cm) for better accuracy of distance estimation. With the full setup we cover more than 220$^\circ$ of the car's surroundings.

The next important sensors are LiDARs. We employ two 3D Velodyne scanners mounted with a slight tilt around the forward-pointing axis of the car as can be seen from the photograph in Fig.~\ref{fig_preview}. The reason is twofold: first, the scanners better cover the area  sideways of the car and second, the rays on the opposite side can measure higher obstacles and do not needlessly scan a roof of the car.

Thermal camera usage is unique among existing datasets for automated driving. The device we employ senses infrared radiation in range of 7.5-13.5~\textmugreek m, which corresponds to peak wavelengths emitted by objects in usual temperatures of -40 to +80 $^\circ$C. The resolution is low in comparison to RGB cameras, but opacity of many objects (e.g. smoke, fog, thin foil) differs for infrared and visible light, so even with 640x512 pixels per frame, the information gain is substantial. The camera if mounted in the most important forward-looking direction.

The fourth kind of sensor mounted on the ATLAS platform is an inertial measurement unit. Besides accelerometers and gyroscopes, the device contains a combined GNSS receiver (GPS, GLONASS, Galileo, BeiDou) and provides additional environmental measurements such as temperature, atmospheric pressure and magnetic field, which have limited usage in automated driving, but we have decided to publish them as well for completeness.

Last but not least, there is a separate combined GNSS receiver to obtain the most precise global localization available. The RTK functionality allows centimetre level accuracy. Additionally, the receiver allows connection of two antennas at the same time allowing to directly obtain a heading vector in global coordinates. This feature was not present in any dataset we know about and can be very valuable as a reference in map building and localization applications. Of course, the receiver provides diagnostics of a reliability of the measurements as well as precise time for other sensors. This feature has a key role in our solution of synchronization and timestamping described in Sec.~\ref{sec_syn_and_stamp}.

\subsection{Calibration}

So far, we have spoken about sensor poses within the ATLAS platform in a somewhat vague manner. The reason comes from difficulties with exact measurements. Each sensor has its own frame of coordinates, whose origin mostly lies within the device and is usually tied with the chassis by a few dimensions with certain tolerance. Although mutual position could be acquired with decent accuracy, the orientation measurement is very sensitive and even a small error could result in faulty alignment of data from multiple sources. Additionally, even if we could measure exactly, there are still manufacturing tolerances, which cannot be dealt with this way.


For this reason, we have decided to perform thorough calibration of sensors. Some of the methods used are suited for estimation of the intrinsic parameters of the device, while others are designed to obtain their mutual pose. Methods used, their settings, calibration data and the best estimates of the desired parameters are provided along the dataset on its website (\url{https://github.com/RoboticsBUT/Brno-Urban-Dataset-Calibrations}). We expect the sensory equipment to change over time and update this material accordingly.

\subsection{Synchronization and Timestamping}
\label{sec_syn_and_stamp}

\begin{figure}[t]
    \centering
    \includegraphics[width=8cm]{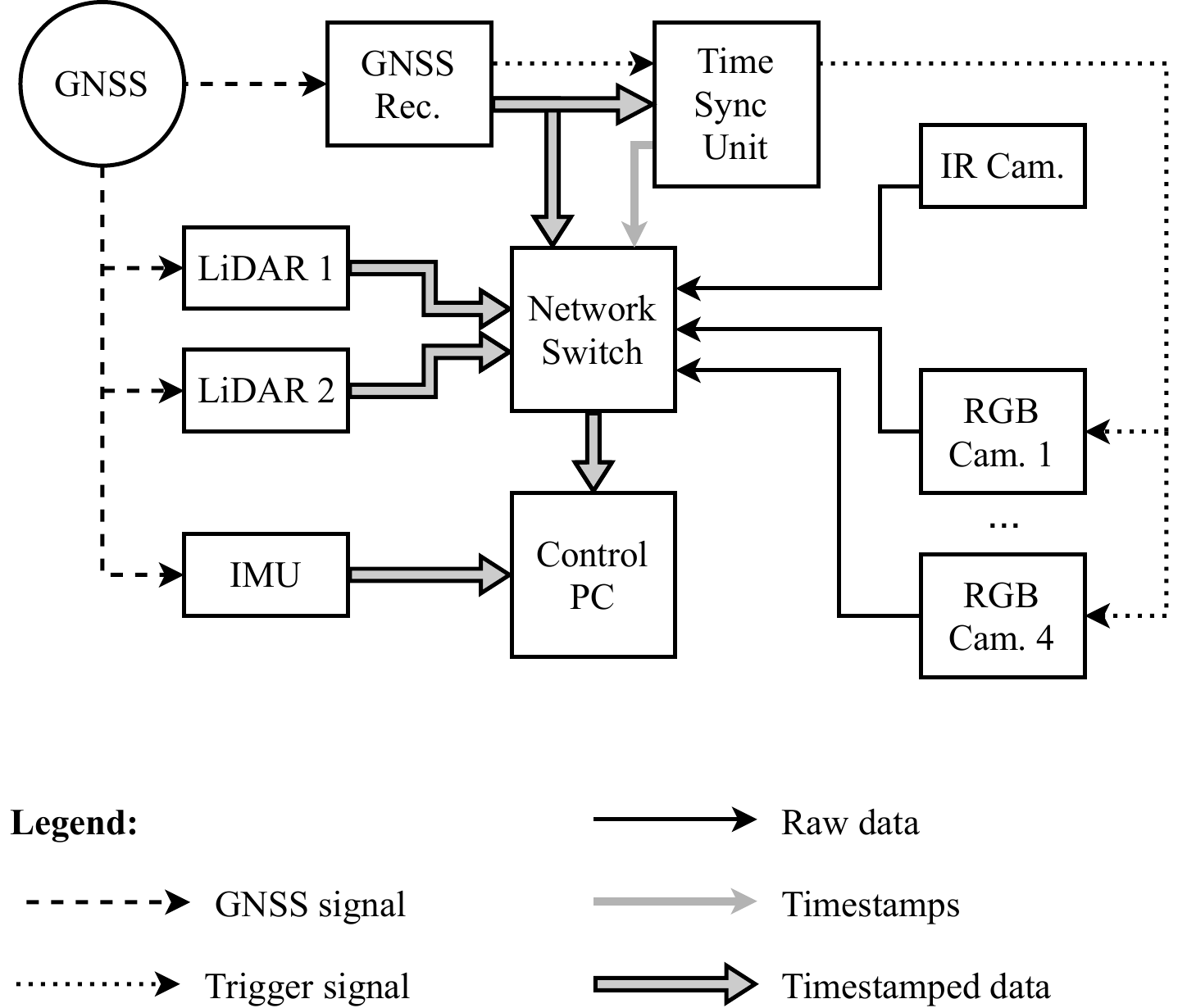}
    \caption{Time synchronization and data-flow diagram of the ATLAS platform.}
    \label{fig_time_sync}
\end{figure}

As already discussed in Sec.~\ref{sec_sota}, exact timing in data acquisition systems is, to some extent, replaceable by dedicated algorithms. We prefer to prevent the problems instead of fix them and the ATLAS platform was built with precise timing in mind. The scheme in Fig.~\ref{fig_time_sync} showing data flow in the system also plots the synchronization and time distribution lanes (dashed arrows). 

The key source of precise time in our system is the GPS signal. Even the most basic receiver needs to maintain time precision in a nanosecond scale to provide usable positioning. Obviously, the Trimble RTK receiver has access to it, but the Velodyne laser scanners and Xsens IMU are equipped with small antennas as well, therefore precise time is available for them directly. For devices with no receiver of the GPS signal, we have designed a synchronization unit, which is clocked by a precise clock source from the Trimble receiver. It can either capture input trigger, pair it with an exact time and send a packet to the recording computer, or generate an output trigger signal with given frequency an send a timestamp corresponding to each firing. The unit is built around a simple microcontroller without operating system or nested interrupts, which allows to maintain transparent timing of its routines and guarantee an upper limit on timestamp-signal mismatch. Taking into account propagation delays in hardware etc., the error is well below 1 ms. With all that precautions, we can completely bypass the system time of the control computer and stamp the data with timing from a trustworthy source.

Currently we use the synchronization unit for triggering of all RGB cameras with common signal. The thermal camera unfortunately can not be directly triggered by an external signal, neither provides an output trigger, but contains precise clock allowing additional corrections. Resulting list of timestamps is than paired with a timestamp sequence from the synchronization unit providing interpolated timing of each frame.
 
An obvious drawback of this approach is a strong dependence on the GPS signal availability. For this reason, all devices drawing the GPS time employ a graceful fallback to local clock source with known accuracy. In the worst case scenario the timing errors break the 1 ms limit after tens of minutes without GPS signal, which is enough to pass tunnels and other problematic locations and regain the exact time. We take a great care not to exceed this limit in any of our recordings.

\section{Dataset}

As mentioned above, the dataset was recorded in Brno, Czech Republic. The ATLAS platform is not yet entirely waterproof, so the range of weather conditions captured is limited. On the other hand, thanks to mid-size of the town, the environmental diversity spans from natural, countryside-looking locations to city center with historical buildings, public transportation and especially large amount of traffic and pedestrians.

\begin{figure*}[t]
    \centering
    \includegraphics[width=17.5cm,height=22.0cm]{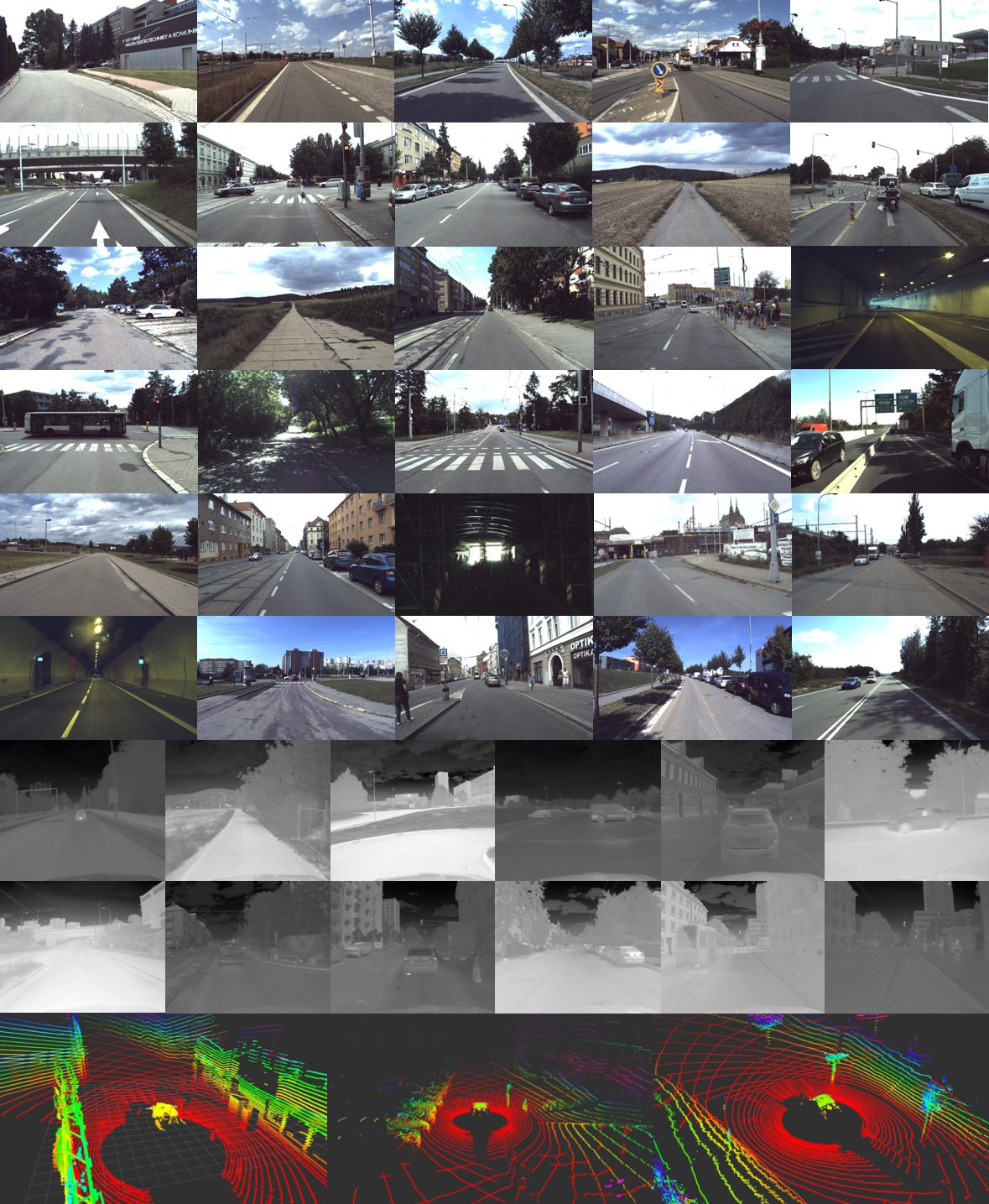}
    \caption{Example of Brno Dataset data visualization. In the first four rows there are the RGB camera images, in the fifth row there are the thermal (IR) camera images, and the LiDAR data visualization on the bottom.}
    \label{fig_data_visualization}
\end{figure*}

\subsection{Content}

It is a good practice to sort the data according to its content. The time of recording serves mostly as a unique identifier and a brief description is good to get a quick overview of the recording, but both are cumbersome to use, if a whole database of all recordings is needed to be searched through. For this reason, we have employed a system of tags, which allow us to highlight the most important content and enable easy filtration of the recordings summarized in Tab.~\ref{tab_tags}.

So far, we have made available 67 recordings of total length of 375.7~km and duration of more than 10 hours. 


\begin{table}[t]
\centering
\begin{tabular}{c|c|c|c|c}
 \hline
 Tag category & Tag & Recordings & Distance & Duration\\
 \hline
 Weather & \makecell{Sunny\\Partly-cloudy} & \makecell{42\\25} & \makecell{245.1\\130.6} & \makecell{6:23\\4:32} \\
\hline
 Daytime & \makecell{Morning\\Noon\\Afternoon\\Evening} & \makecell{15\\26\\21\\5} & \makecell{60.1\\175.6\\96.4\\43.6} & \makecell{1:48\\4:04\\3:37\\1:24} \\
 \hline
 Environment & \makecell{City\\Suburb\\Country\\Highway} & \makecell{36\\21\\6\\4} & \makecell{181.9\\71.0\\48.1\\74.7} & \makecell{5:56\\2:34\\1:16\\1:08} \\
 \hline
\end{tabular}
\caption{Statistics of the dataset content. Length is measured in kilometers and duration in hours:minutes format.}
\label{tab_tags}
\end{table}

\subsection{Data Structure}

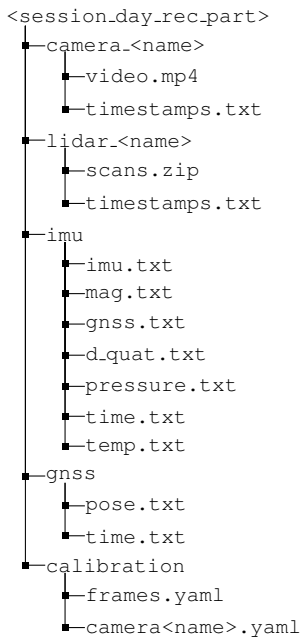
\begin{figure}[t]
    \centering

{\footnotesize    
\begin{forest}
  for tree={
    font=\ttfamily,
    inner sep=0pt,
    l=0pt,
    l sep=2pt,
    grow'=0,
    child anchor=west,
    parent anchor=south,
    anchor=west,
    calign=first,
    edge path={
      \noexpand\path [draw, \forestoption{edge}]
      (!u.south west) +(7.5pt,0) |- node[fill,inner sep=1.25pt] {} (.child anchor)\forestoption{edge label};
    },
    before typesetting nodes={
      if n=1
        {insert before={[,phantom]}}
        {}
    },
    fit=band,
    before computing xy={l=15pt},
  }
[<session\_day\_rec\_part>
  [camera\_<name>
    [video.mp4]
    [timestamps.txt]
  ]
  [lidar\_<name>
    [scans.zip]
    [timestamps.txt]
  ]
  [imu
    [imu.txt]
    [mag.txt]
    [gnss.txt]
    [d\_quat.txt]
    [pressure.txt]
    [time.txt]
    [temp.txt]
  ]
  [gnss
    [pose.txt]
    [time.txt]
  ]
  [calibration
    [frames.yaml]
    [camera<name>.yaml]
  ]
]
\end{forest}}
\caption{Data structure of a recording with separated folders for every sensor and common data.}
    \label{fig_data_struct}
\end{figure}

The structure of a single recording follows the scheme in Fig.~\ref{fig_data_struct}. RGB and thermal camera data are distributed as H265 video and the LiDAR scans are compressed into .zip archives to reduce their size as much as possible. Other data such as timestamps and calibration files occupy a negligible amount of memory and are stored in human readable .txt and .yaml files.

\begin{figure}[t]
    \centering
    \includegraphics[width=8.5cm,height=4.8cm]{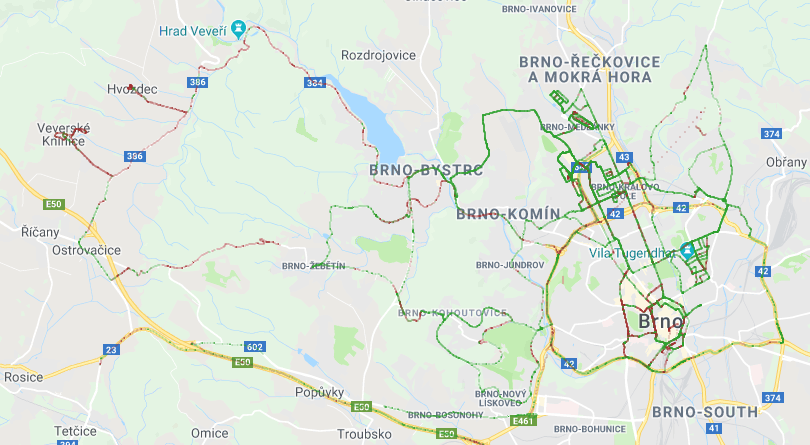}
    \caption{Map visualization of the roads traveled during the dataset recording. Green - RTK positioning + heading, Red - non RTK positioning or missing the heading data. Brno coordinates: 49.2002211N, 16.6078411E.}
    \label{fig_data_gps_map}
\end{figure}

\subsection{Software and Development Tools}

The recording session runs fully on the Robot Operation System (ROS). This allows us to create highly scalable solution which is compatible with many other projects using ROS backend as a base line for development of robotics applications. To satisfy wider audience and more comfortable usage, we publish the data in an easily readable raw format and provide a script for conversion into the ROS bag. 

We also provide a set of several Python-OpenCV based scripts that helps to process the video data into separated frame files, or the drawing the trajectory into the Google Maps, etc. The software is available from \url{https://github.com/RoboticsBUT/Brno-Urban-Dataset-Tools}, or as a submodule of the dataset git repository in the \texttt{tools/} folder.

\section{Conclusion and Future Work}

In this paper, we have presented a new dataset for autonomous driving research recorded in Brno - Czech Republic. We provide state of the art sensory measurements with three key additions exceeding other datasets available. First, most routes where GNSS signal was available are accompanied with centimeter accurate global position and heading from differential GNSS receiver. Second important feature is synchronization and timestamping of the data with sub-millisecond precision allowing simpler data fusion and evaluation of the algorithms, where temporal shifts in measurements are part of the stochastic model. The last addition is the infrared camera significantly increasing detection and recognition capabilities of the ATLAS measurement platform. 

At the time of writing of this paper 67 recordings with total length of 375.7~km and duration of more than 10 hours are available. The recordings are tagged with respect to the environment, weather and other events encountered and provided in easily usable format through the project page: \url{https://github.com/RoboticsBUT/Brno-Urban-Dataset}. Data collection is a long term process and we expect the dataset to grow over time with various new recording required by our research. We plan to densely cover a smaller region in Brno for map-building applications and to obtain more data acquired with problematic lightning and weather conditions or the winter sessions. We are also considering waterproofing of the ATLAS platform to cover full range of weather conditions encountered in middle Europe.

\bibliography{texts/main.bib}
\bibliographystyle{IEEEtran}

\end{document}